\documentclass[11pt]{article}

\usepackage[preprint]{acl}

\usepackage{times}
\usepackage{latexsym}

\usepackage[T1]{fontenc}

\usepackage[utf8]{inputenc}

\usepackage{microtype}

\usepackage{inconsolata}

\usepackage{graphicx}
\usepackage{comment}
\usepackage{enumitem}  
\usepackage{booktabs}
\usepackage{multirow}
\usepackage{subcaption}
\usepackage{amsmath}
\usepackage{tcolorbox}

\usepackage[table,rgb]{xcolor} 
\definecolor{seafoamgreen}{HTML}{66c2a4}
\definecolor{mylightorange}{HTML}{fc8d62}
\definecolor{mylightblue}{HTML}{8ea0cb}
\colorlet{seafoamgreen_trans}{seafoamgreen!50!white}
\colorlet{mylightorange_trans}{mylightorange!50!white}
\colorlet{mylightblue_trans}{mylightblue!50!white}

\usepackage{tcolorbox}
\tcbuselibrary{breakable}
\usepackage{fvextra} 

\DefineVerbatimEnvironment{PromptVerb}{Verbatim}{
  fontsize=\scriptsize,
  breaklines=true,
  breakanywhere=true,
  breaksymbolleft={}
}
\usepackage{pifont}
\newcommand{\cmark}{\ding{51}}%
\newcommand{\xmark}{\ding{55}}%

\usepackage[dvipsnames]{xcolor}
\usepackage{xcolor}
\definecolor{DarkGreen}{RGB}{0,90,0}   
\newcommand{\GreenSign}{\textcolor{DarkGreen}{\cmark}}
\definecolor{DarkerRed}{RGB}{170,0,0}
\newcommand{\RedSign}{\textcolor{DarkerRed}{\xmark}}

\newcommand{\GenFig}{\textsc{GenFig1}}
\title{\GenFig: Visual Summaries of Scholarly Work\\ as a Challenge for Vision-Language Models}

\author{Yaohan Guan, Pristina Wang \\ 
\bf Najim Dehak, Alan Yuille, Jieneng Chen, Daniel Khashabi \\
  Johns Hopkins University  \\
  Correspondence: \texttt{yguan19@jhu.edu, pwang71@alumni.jh.edu} 
  }

\begin{document}
\maketitle
\begin{abstract}
In many science papers, ``Figure 1'' 
serves as the primary visual summary of the core research idea.
  These figures are visually simple yet conceptually rich, often requiring significant effort and iteration by human authors to get right, highlighting the difficulty of science visual communication. With this intuition, we introduce \GenFig, a benchmark for generative AI models (\emph{e.g.}, Vision–Language Models). 
  \GenFig{} evaluates models for their ability to produce figures that clearly express and motivate the central idea of a paper (title, abstract, introduction, and figure caption) as input.
  Solving \GenFig{} 
  requires more than producing visually appealing graphics: 
  the task entails reasoning for text-to-image generation that couples scientific understanding with visual synthesis. Specifically, models must \textit{(i)} comprehend and grasp the technical concepts of the paper, \textit{(ii)} identify the most salient ones, and \textit{(iii)} design a coherent and aesthetically effective graphic 
  that conveys those concepts visually and is faithful to the input. 
    We curate the benchmark from papers published at top deep-learning conferences, apply stringent quality control, and introduce an automatic evaluation metric that correlates well with expert human judgments. We evaluate a suite of representative models on \GenFig{} and demonstrate that the task presents significant challenges, even for the best-performing systems.
    We hope this benchmark serves as a foundation for future progress in multimodal AI.\footnote{We release our data at \url{https://huggingface.co/datasets/yaohanguan/GenFig1}}
\end{abstract}

\begin{figure*}
    \centering
    \includegraphics[width=0.99\linewidth,trim=0cm 2cm 0cm 0cm]{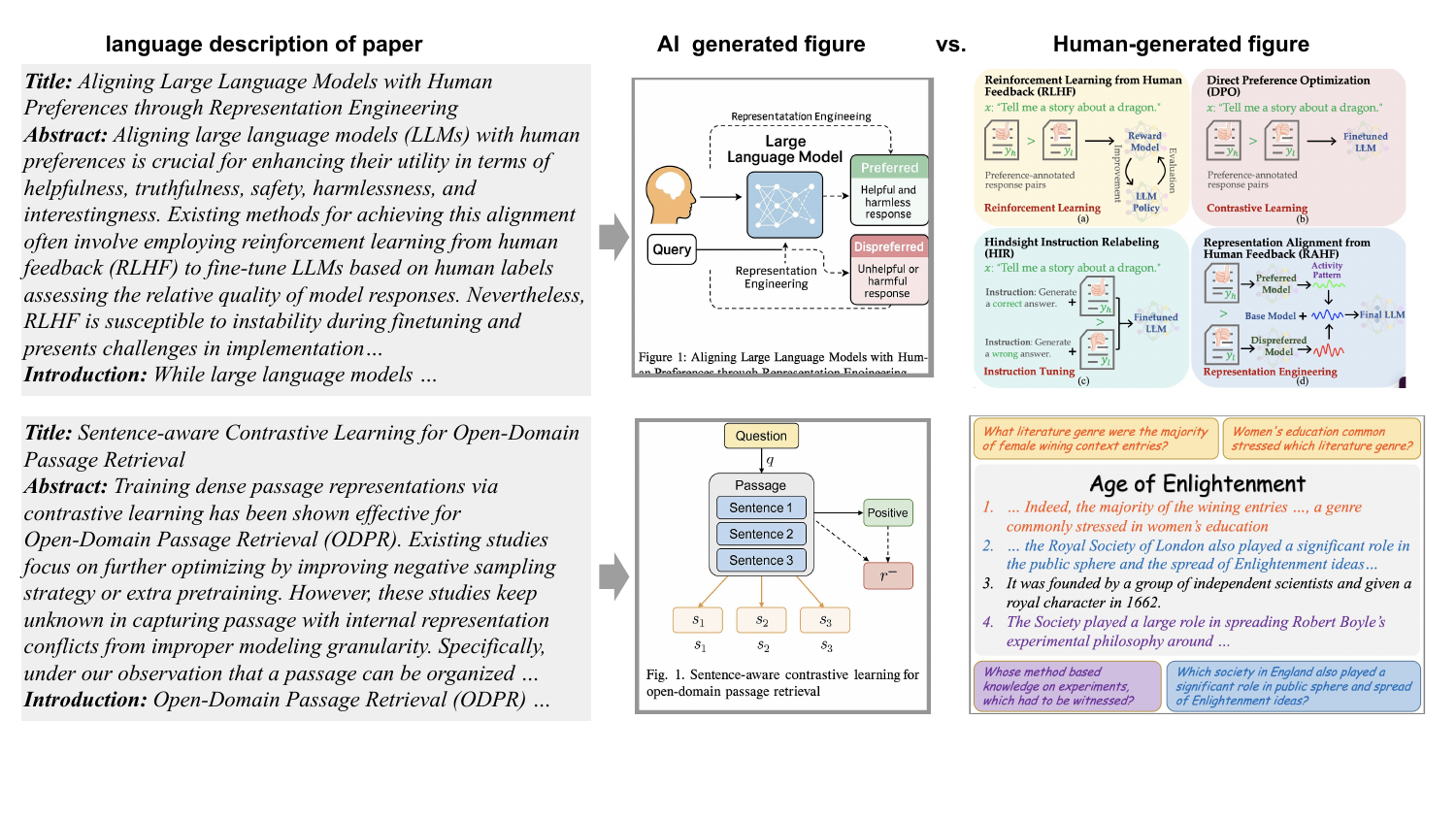}
    \caption{
    Examples from \GenFig (the first row from \cite{liu2024aligning} and second row from \cite{wu2022sentence}).
    The task is to produce figures that clearly express and motivate the central idea of a paper (title, abstract, introduction, and figure caption) as input.
    Example responses from models we evaluate are shown in the middle column. 
    Solving \GenFig{} requires more than just visually appealing graphics: the task entails cross-modal reasoning that couples scientific understanding with visual synthesis.
    }
    \label{fig:teaser-figure}
\end{figure*}
\section{Introduction}
Scientific figures are central to  visual communication in research, play a crucial role in how ideas are communicated, understood, and remembered.
In many papers—especially in disciplines like AI, NLP and computer vision, ``Figure 1'' often functions as the conceptual centerpiece: a visual distillation of the paper's core motivation and contribution. These figures are not mere illustrations; they are designed to be both visually simple and conceptually rich, serving as intuitive gateways into complex technical ideas. Compared with later figures, which more often elaborate specific components, implementation details, or experimental analyses, Figure 1 is typically designed to orient the reader and foreground the paper’s central contribution. Creating such figures is a non-trivial task, typically requiring significant domain understanding, design iteration, and careful abstraction by authors. This highlights a broader challenge: 
\textit{conveying scientific ideas visually is inherently difficult, especially when abstract concepts must be conveyed succinctly and clearly}.

This challenge presents a unique opportunity for generative AI systems. 
Notably, recent advances in multimodal generative AI systems powered by vision–language models (VLMs)~\cite{hurst2024gpt}
have achieved remarkable progress, particularly in generating photorealistic images and even videos from natural language prompts~\cite{singermake}. 
Yet it remains unclear whether these models can perform perform the reasoning required for text-to-diagram generation. 
In particular, can these models digest technical content (e.g., paper title, abstract, introduction, and figure caption) 
and 
synthesize visuals that are both aesthetically coherent and conceptually faithful to the underlying research idea?

To investigate this question, we introduce \GenFig{}, a benchmark designed to evaluate the ability of generative AI models to produce ``Figure 1''-style scientific illustrations. Unlike conventional image generation tasks, \GenFig{} requires models to couple scientific understanding with visual synthesis: they must (i) comprehend the content of a paper, (ii) identify its key contributions, and (iii) design a figure that visually communicates these ideas with clarity and fidelity.

We evaluate a range of representative VLMs on \GenFig{}. This includes various combinations of VLMs to interpret and to translate reasoning about scientific content into effective visual layouts. We find that even the strongest models struggle with the task. The figures they generate often lack conceptual grounding, visual coherence, and faithful alignment with the source material—indicating that current models fall short of the depth of reasoning required for high-quality scientific visualization.

Furthermore, we evaluate a range of automatic evaluation metrics to measure different aspects of figure quality, such as clarity, aesthetic quality, etc.  
We measure the correlation of these metrics with domain expert human judgments to identify the most reliable metrics, enabling scalable and consistent benchmarking. Our experiments show that the VLM-as-a-Judge metric, especially on information density, exhibits the strongest correlation with expert human judgments.

We hope that \GenFig{} serves as a foundation for future progress in reasoning-driven  multimodal systems, \textit{science of AI}. At the same time, it supports downstream applications in \textit{AI for science}, by pushing toward tools that can assist researchers in communicating their ideas more effectively.

\begin{table*}[t]
    \setlength{\tabcolsep}{3pt}
    \centering
    \small
    \resizebox{1\textwidth}{!}{
    \begin{tabular}{lcccccc} 
    \toprule
    \multirow{ 2}{*}{Dataset}
    & \multicolumn{3}{c}{Paper Contents Included}  & \multirow{ 2}{*}{Source}  & \multirow{ 2}{*}{\shortstack{Excluded \\Quant. Figures}} \\
    \cmidrule(lr){2-4}
    & Figure & Caption & Paper Text &  & \\
    \midrule
FigureSeer~\cite{Siegel2016FigureSeerPR}     & \GreenSign & \RedSign & \RedSign & CiteSeerX & \RedSign\\
DVQA~\cite{kafle2018dvqaunderstandingdatavisualizations}     & \GreenSign & \RedSign & \RedSign & synthetic &  \RedSign\\
    FigureQA~\cite{kahou2018figureqaannotatedfiguredataset}     & \GreenSign & \RedSign & \RedSign & synthetic &  \RedSign\\

FigCAP~\cite{chen2019figurecaptioningreasoningsequencelevel}     & \GreenSign & \GreenSign & \RedSign & synthetic & \RedSign\\
SciCap~\cite{hsu2021scicapgeneratingcaptionsscientific}     & \GreenSign & \GreenSign & \RedSign & arXiv &   \RedSign \\
Paper2Fig100k~\cite{rodriguez2022ocrvqgantamingtextwithinimagegeneration}     & \GreenSign & \GreenSign & \RedSign & arXiv &  \RedSign\\
SciCap+~\cite{yang2023scicapknowledgeaugmenteddataset}     & \GreenSign & \GreenSign & text context of figures & arXiv & \RedSign\\
SciFIBench~\cite{roberts2024scifibenchbenchmarkinglargemultimodal}     & \GreenSign & \GreenSign & \RedSign & arXiv & \RedSign\\
CharXiv~\cite{wang2024charxivchartinggapsrealistic}     & \GreenSign & \RedSign & \RedSign & arXiv & \RedSign\\
ArXivCap~\cite{li2024multimodalarxivdatasetimproving}     & \GreenSign & \GreenSign & \RedSign & arXiv & \RedSign\\
M-Paper~\cite{hu2024mplugpaperowlscientificdiagramanalysis}     & \GreenSign & \GreenSign & text context of figures & PaperWithCode & \RedSign\\
MMSci~\cite{li2025mmscidatasetgraduatelevelmultidiscipline}     & \GreenSign & \GreenSign & full text & Nature Communications & \RedSign\\
SridBench ~\cite{chang2025sridbench}  & \GreenSign & \GreenSign & related section &  arXiv and Nature & \RedSign\\

FigureBench (\citealp{zhu2026autofigure}) & \GreenSign & \RedSign & text context of figures&
Research-14K, arXiv, blogs, OpenStax &  \GreenSign \\

PaperBanana \cite{zhu2026paperbananaautomatingacademicillustration} & \GreenSign & \GreenSign & methodology section &   NeurIPS 2025 & \GreenSign\\
     \midrule
    \textbf{\GenFig{}} (ours)     & \GreenSign & \GreenSign & \textbf{intro + abstract} & \textbf{top AI/ML venues} &  \GreenSign\\
    \bottomrule
    \end{tabular}
    }
    \caption{Comparison of our dataset with previous datasets. Unlike previous datasets, ours includes a broad collection of peer-reviewed papers on top AI/ML venues, with a targeted focus on figure with caption, coupled with the introductions and abstracts.
    }
    \label{tab:dataset_comparison}
\end{table*}

\section{Related Work}

We discuss related work in two broad categories: benchmarks and figure generation frameworks. 
\vspace{-0.5\baselineskip}
\paragraph{Related datasets:}
We compare all the previous datasets in Table~\ref{tab:dataset_comparison}.
Several datasets target related but distinct tasks which we briefly highlight. 
Some datasets are motivated by question answering (QA) on scientific figures. For example, FigureQA~\cite{kahou2018figureqaannotatedfiguredataset} and DVQA~\cite{kafle2018dvqaunderstandingdatavisualizations} are two datasets with questions and answers about common scientific-style plots. 
These datasets rely on synthetic figures, which lack the richness of natural researcher-created figures.
There are resources that are motivated by other aspects of figure analysis. For instance,  few datasets~\cite{hu2024mplugpaperowlscientificdiagramanalysis,roberts2024scifibenchbenchmarkinglargemultimodal,chen2019figurecaptioningreasoningsequencelevel,hsu2021scicapgeneratingcaptionsscientific} were developed to facilitate multimodal diagram captioning and interpretation. 
CharXiv~\cite{wang2024charxivchartinggapsrealistic} is a chart understanding benchmark extracted from arXiv. 
ArXivCap is a figure-caption dataset extracted from arXiv by ~\citet{li2024multimodalarxivdatasetimproving}.
FigureSeer dataset is a dataset created by ~\citet{Siegel2016FigureSeerPR} extracted from CiteSeerX papers to facilitate figure parsing. 
FigureSeer dataset contains figures and figure annotations. \citet{rodriguez2022ocrvqgantamingtextwithinimagegeneration} created Paper2Fig100k to facilitate figure generation. Paper2Fig100k is a dataset with OCR results, aspect ratios of images, paper figures and their captions extracted from arXiv. More recently, \citet{chang2025sridbench} proposed SridBench for figure generation, which includes images, captions, and related sections. \citet{zhu2026autofigure} introduced FigureBench, a benchmark of 3,300 long-form scientific text--figure pairs drawn from scientific papers, surveys, blogs, and textbooks, where selected figures are paired with source text describing their key visual elements. \cite{zhu2026paperbananaautomatingacademicillustration} introduced PaperBananaBench, a benchmark for methodology-diagram generation curated from NeurIPS 2025 publications, containing 584 samples in total, each consisting of a methodology description, a methodology diagram, and its caption; these are split into a 292-example test set and a 292-example reference set.
Our proposed dataset, \GenFig{}, introduces aspects not covered by existing datasets (Table~\ref{tab:dataset_comparison}). Specifically,  we focus on generating illustrative figures (i.e., ``Figure 1'') of peer-reviewed scientific papers based on their textual content (e.g., abstract, introduction).

\paragraph{Scientific figure generation:}
The broader field of image generation has rapidly advanced over the past decade, evolving from GANs~\cite{goodfellow2020generative} to diffusion models~\cite{sohl2015deep, ho2020denoising, rombach2022high}. Diffusion models have emerged as a leading class of generative models, achieving state-of-the-art sample quality in tasks such as text-to-image generation~\cite{GLIDE, rombach2022high, Imagen} and image-to-image translation~\cite{SR3, Repaint, DDRM}. ControlNet~\cite{zhang2023adding} extended latent diffusion models~\cite{rombach2022high} with dense conditioning inputs (e.g., semantic maps), enabling precise layout-guided image generation~\cite{zheng2023layoutdiffusion, inoue2023layoutdm}. These advances lay a strong foundation for automating the creation of scientific figures.
Within the scientific figure generation community, efforts have expanded beyond diffusion models—though they remain a core component. In particular, code generation approaches using languages like TikZ and SVG have become increasingly prominent. For instance, \citet{belouadi2024automatikz} focused on generating scientific figures via TikZ code, while \citet{rodriguez2024starvectorgeneratingscalablevector} explored both text-to-SVG and image-to-SVG generation. \citet{rodriguez2023figgentextscientificfigure} introduced FigGen, a diffusion-based method for generating scientific figures from natural language.
In our work, we benchmark a range of baselines to evaluate the effectiveness of these figure generation approaches (\S\ref{sec:evaluation}), including standalone diffusion models, hybrid methods that integrate language reasoning, and code generation–based pipelines.


\section{\GenFig}
The construction pipeline of our dataset contains three stages: data collection, filtering, and analysis.


\subsection{Data Curation}
\label{subsec:curation}

\paragraph{Initial data collection:}
We identify top-tier AI venues across multiple domains: Computer Vision (CVPR), Natural Language Processing (ACL, NAACL and EMNLP), and general AI/Machine Learning (ICLR, NeurIPS) from 2020 through 2025. Our collection excludes NAACL for years when it did not occur (2020, 2023, and 2025), as well as ACL, EMNLP and NeurIPS 2025, which have not yet taken place when we crawled the data. We crawled the ACL, NAACL, EMNLP long/main track paper titles from ACL Anthology's website. We crawled older CVPR accepted papers (older than 2022) from open-access and  newer accepted papers (newer than 2022) from cvpr.thecvf.com and ICLR accepted (poster/spotlight/oral) papers using OpenReview API. We crawled NeurIPS accepted papers from \url{papers.nips.cc}.

\paragraph{Mapping papers to their \LaTeX{} source:}
After retrieving  paper titles, we used the arXiv API to search for arXiv paper IDs with the titles and created a database that maps paper titles to arXiv IDs. arXiv API search are not always successful, so we don't have all of the arXiv IDs for all crawled paper titles. Using the retrieved arXiv IDs, we download the paper \LaTeX{} source from arXiv and save them.

\paragraph{Extracting content from \LaTeX{} source:}
With all the \LaTeX{} source, we use \texttt{plasTeX}~\cite{plastex} to parse the \LaTeX{} documents into nested \texttt{plasTeX} document objects. With the nested \texttt{plasTeX} document objects and regular expression, we are able to extract abstract section, introduction section, the section of the first figure section in the paper. We then extract the \LaTeX{} text source from these sections and use \texttt{PyLaTeX}~\cite{PyLaTeX} to parse \texttt{LaTeX} macros and math formulas into plain text. Lastly, we extract the image path of the figure image using regular expression. 
Because parsing is computationally expensive,  we used \texttt{Pebble}~\cite{pebble} to run all extractions in parallel.

\subsection{Dataset Cleaning and Filtering}
\label{subsec:cleaning}
Despite a high-quality extraction pipeline, some figures were incorrectly extracted and required filtering. We removed icons (images under 100 pixels) and black images, as well as figures consisting of quantitative results (e.g., line charts, histograms, box plots, heatmaps) focused on model performance or parameter settings. Since our task inputs lack experimental data, reproducing such figures is not feasible. Unlike prior datasets (see Table~\ref{tab:dataset_comparison}, last column), we apply this filtering to focus on figures that act as visual blueprints rather than those reporting raw statistics.

\subsection{Data Analysis}
\label{subsec:data:analysis}

\paragraph{The resulting dataset:}

Our final dataset consists of tuples of (Title, Abstract, Introduction, Figure 1 caption, Figure 1) from 9,976 research papers that we collected from selected conferences in three domains: AI/ML, Computer Vision (CV), and Natural Language Processing (NLP).

Within the dataset, AI/ML papers make up the largest proportion (44.7\%), followed by CV (31.6\%) and NLP (23.7\%) (see Table~\ref{tab:domain_breakdown}). Among the NLP conferences, EMNLP contributes the most papers, while NeurIPS is the largest among the AI/ML venues. 
Table \ref{tab:venue_counts} in the Appendix details the number of papers retained after filtering for each venue by year from 2020 to 2025.

\begin{table}[ht]
  \setlength{\tabcolsep}{3pt}
  \small
  \centering
  \resizebox{0.48\textwidth}{!}{
  \begin{tabular}{llccc}
    \toprule
    Domain & Venue & \# Papers & Domain Share (\%) & Total Share (\%) \\
    \midrule
    \multirow{3}{*}{NLP} 
      & ACL   & 686   & 29.0 & 6.9 \\
      & NAACL & 199   & 8.4  & 2.0 \\
      & EMNLP & 1,484 & 62.6 & 14.9 \\
    \cmidrule(l){2-5}
      & \textbf{Subtotal} & \textbf{2,369} & \textbf{100.0} & \textbf{23.7} \\
    \midrule
    CV & CVPR & \textbf{3,148} & \textbf{100.0} & \textbf{31.6} \\
    \midrule
    \multirow{2}{*}{AI/ML} 
      & ICLR    & 2,121 & 47.6 & 21.3 \\
      & NeurIPS & 2,338 & 52.4 & 23.4 \\
    \cmidrule(l){2-5}
      & \textbf{Subtotal} & \textbf{4,459} & \textbf{100.0} & \textbf{44.7} \\
    \midrule
    \textbf{Total} & & \textbf{9,976} & & \textbf{100.0} \\
    \bottomrule
  \end{tabular}
  }
  \caption{Distribution of papers in our \GenFig{} by domain and venue (2020--2025). It covers six major venues across three research domains, with AI/ML (NeurIPS, ICLR) contributing nearly half of the papers, followed by  CV (CVPR) and NLP (EMNLP, ACL, NAACL).}
  \label{tab:domain_breakdown}
\end{table}

\paragraph{Taxonomy:}

Figure 1 generation is a text-conditioned generation task. We assume with more specific text conditions, the generated image space becomes narrower. Although the provided textual input is already targeted, introducing a taxonomy category label can further restrict the image generation space and enable the model to generate images in a more condition-aware way. To better understand the taxonomy of Figure 1s, we sample 50 Figure 1s from the dataset and manually categorize them into three main taxonomies—\textbf{Overview}, \textbf{Example}, and \textbf{Experimental Results}—as well as finer-grained subcategories described below.

\begin{itemize}
[leftmargin=*,itemsep=0pt]
\item \colorbox{seafoamgreen_trans}{\textbf{Overview}}. The Overview category covers figures that depict the overall structure or workflow of the proposed model or method. It includes two main subcategories:
    \begin{itemize}[leftmargin=*, itemsep=0pt, topsep=0pt]
        \item \emph{Model Architecture}:  providing a detailed visualization of the model’s structural components and their interconnections.
        \item \emph{Method}: illustrating the high-level workflow or procedural framework of the method, including pipelines or algorithmic steps. These may include frameworks that outline the end-to-end process, often with step-by-step textual examples, as well as algorithmic diagrams—such as those for EM algorithms—that explicate each step of the procedure.
    \end{itemize}

\item \colorbox{mylightorange_trans}{\textbf{Example}}.
Figures in this category illustrate specific instances for clarity and context. It contains three subcategories:
    \begin{itemize}[leftmargin=*, itemsep=0pt, topsep=0pt]
        \item \emph{Background}: introducing or contextualizing the research problem or the task being studied.
        \item \emph{Method}: unlike the high-level method in the Overview category, figures here focus specifically on concrete input-output examples or demonstrate comparisons of model outputs against baseline methods.
        \item \emph{Dataset}: visualizing representative samples from datasets or the structure and composition of datasets. 

    \end{itemize}

\item  \colorbox{mylightblue_trans}{\textbf{Experimental Results}}. 
This category includes Figure~1s that present or compare experimental findings. We exclude them because they typically do not convey a paper’s core concepts at first glance, making them less effective as Figure~1s.
    \begin{itemize}[leftmargin=*, itemsep=0pt, topsep=0pt]
      \item \emph{Model Metrics}: numerical performance metrics specific to a single method (e.g., average number of experts per layer after pruning).
      \item \emph{Comparative Results}: side-by-side performance comparisons among multiple methods.
      \item \emph{Descriptive Analysis}: visualizations of broader trends or phenomena in research(e.g., annual publication trends in text simplification), or the data distribution. 
    \end{itemize}

\end{itemize}

\begin{figure}[ht]
    \centering
        \includegraphics[width=1\linewidth]{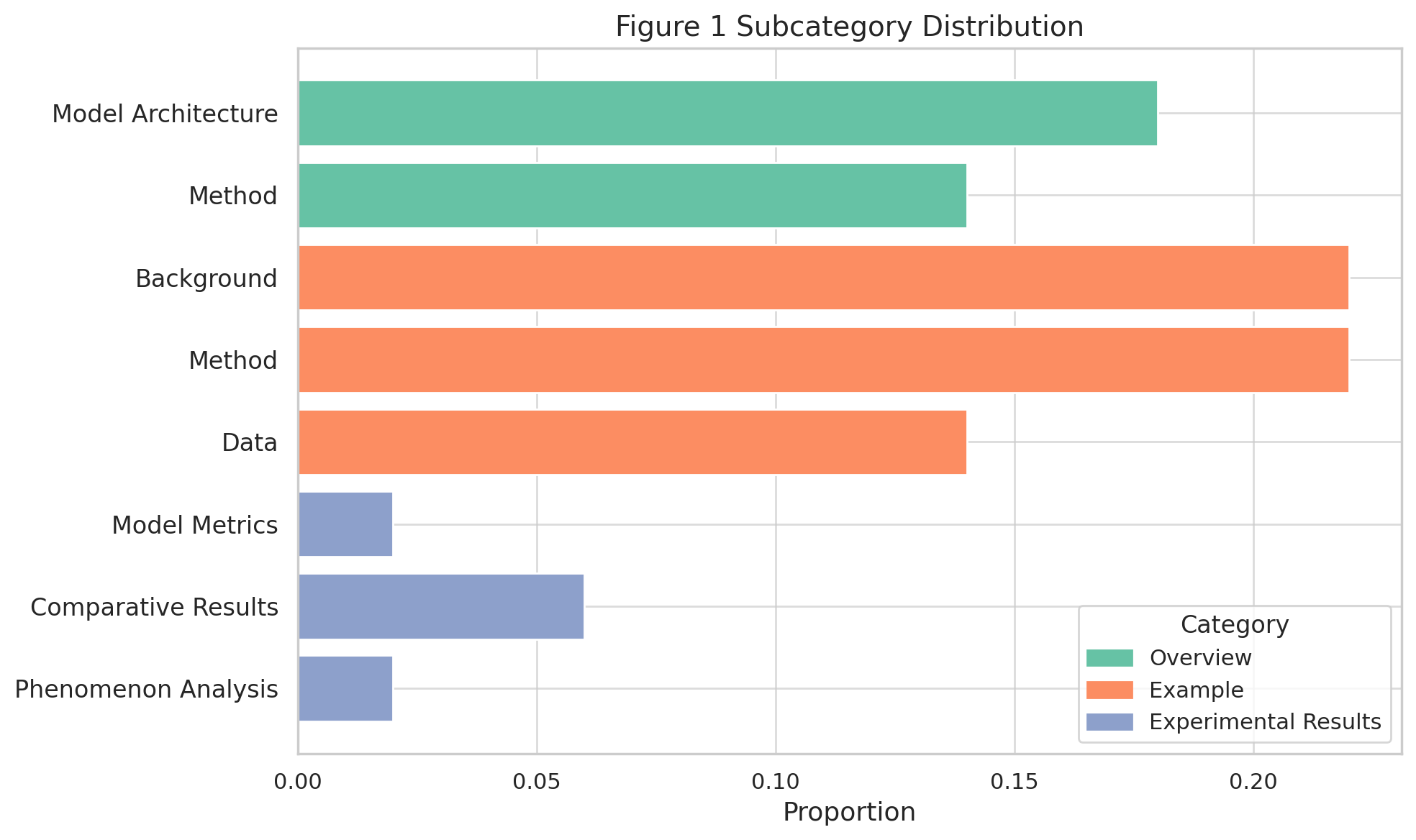}
    \caption{Resulted taxonomy of Figure 1s. We define three taxonomies(Overview, Example, and Experimental Results) and multiple sub-taxonomies, where Overview and Example contribute more. Among them, Example–Background and Example–Method are the most frequent, followed by Overview–Model Architecture and Overview–Method.}
    \label{fig:Taxonomy.png}
\end{figure}

In Figure \ref{fig:Taxonomy.png}, we can see that Overview and Example contribute more. Among them, Example–Background and Example–Method are the most frequent, followed by Overview–Model Architecture and Overview–Method.

\paragraph{Filtering effect and coverage}
Table \ref{tab:venue_filter_comparison} in the Appendix provides a detailed comparison between the raw corpus sizes and the final dataset after filtering, as well as the corresponding filtering rates for each venue and year. Papers with only quantitative experiment results are filtered out. Filtering rates generally range from 66\% to over 93\% depending on the venue and year. CVPR consistently exhibits the highest retention rates, exceeding 90\% in most years. This filtering ensures high data quality and consistency across venues, supporting reliable benchmarking.

\begin{figure*}[th]
    \vspace{2mm}
    \centering

    \begin{subfigure}[b]{0.3\textwidth}
        \centering
        \includegraphics[width=\linewidth,height = 4cm]{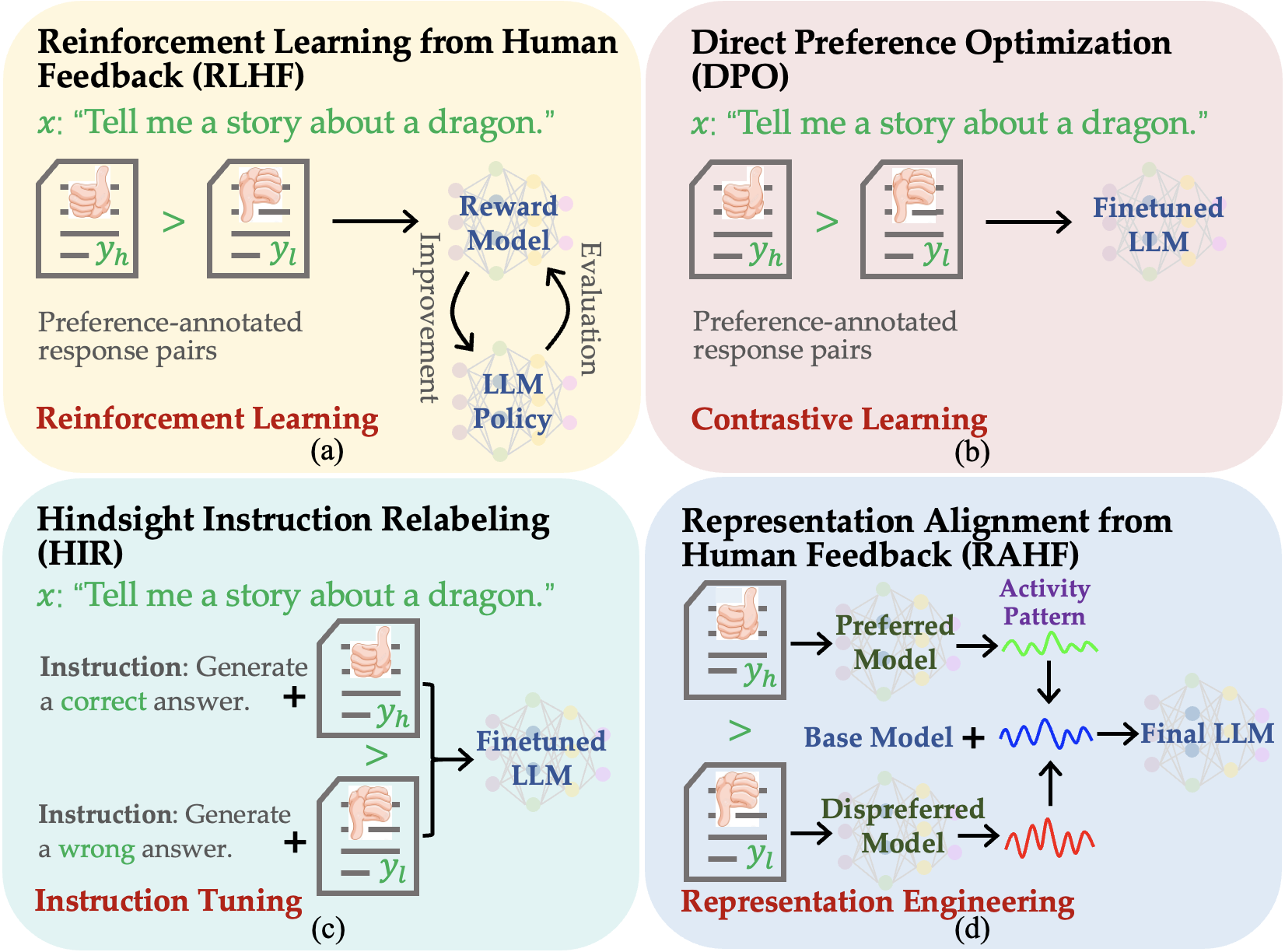}
        \caption{Humans}
        \label{fig:sub1}
    \end{subfigure}
    \hfill
    \begin{subfigure}[b]{0.3\textwidth}
        \centering
        \includegraphics[width=\linewidth,height = 4cm]{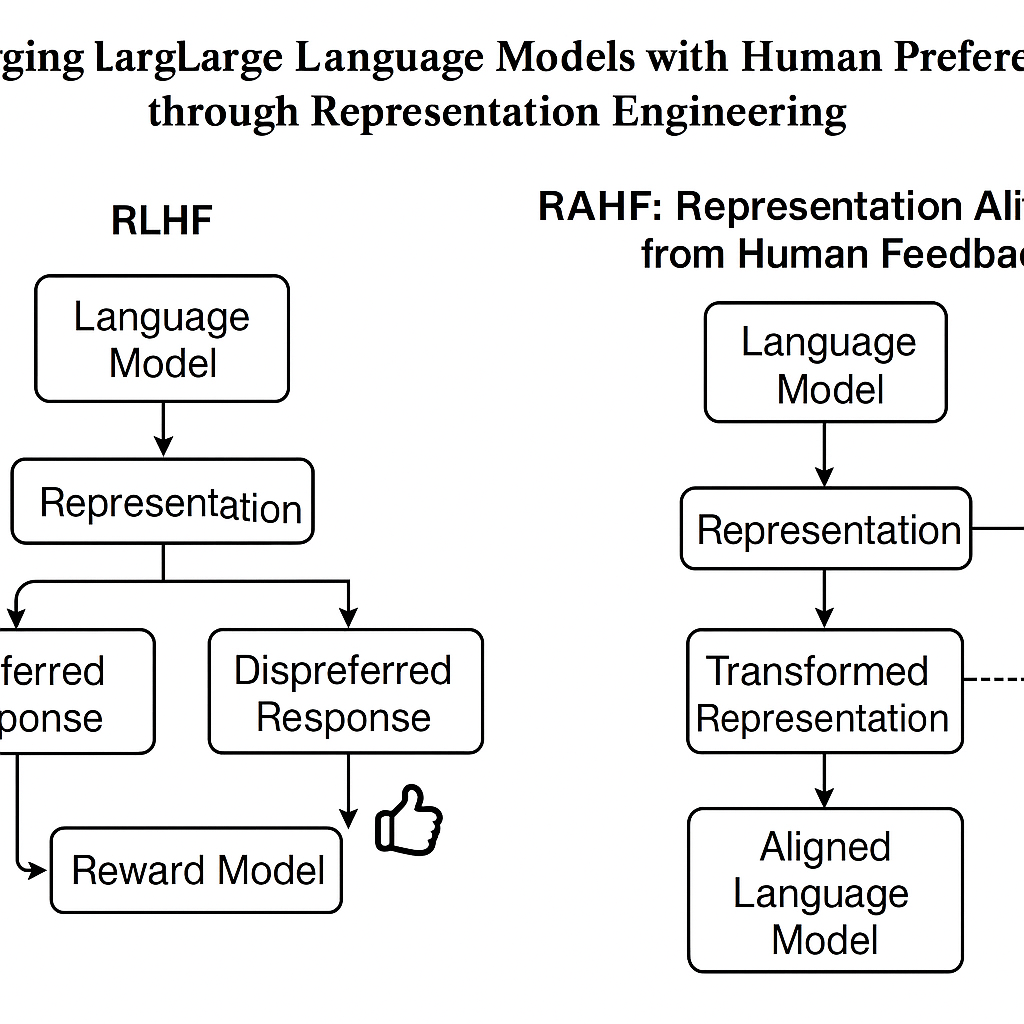}
        \caption{Zero-shot}
        \label{fig:sub2}
    \end{subfigure}
    \hfill
    \begin{subfigure}[b]{0.3\textwidth}
        \centering
        \includegraphics[width=\linewidth,height = 4cm]{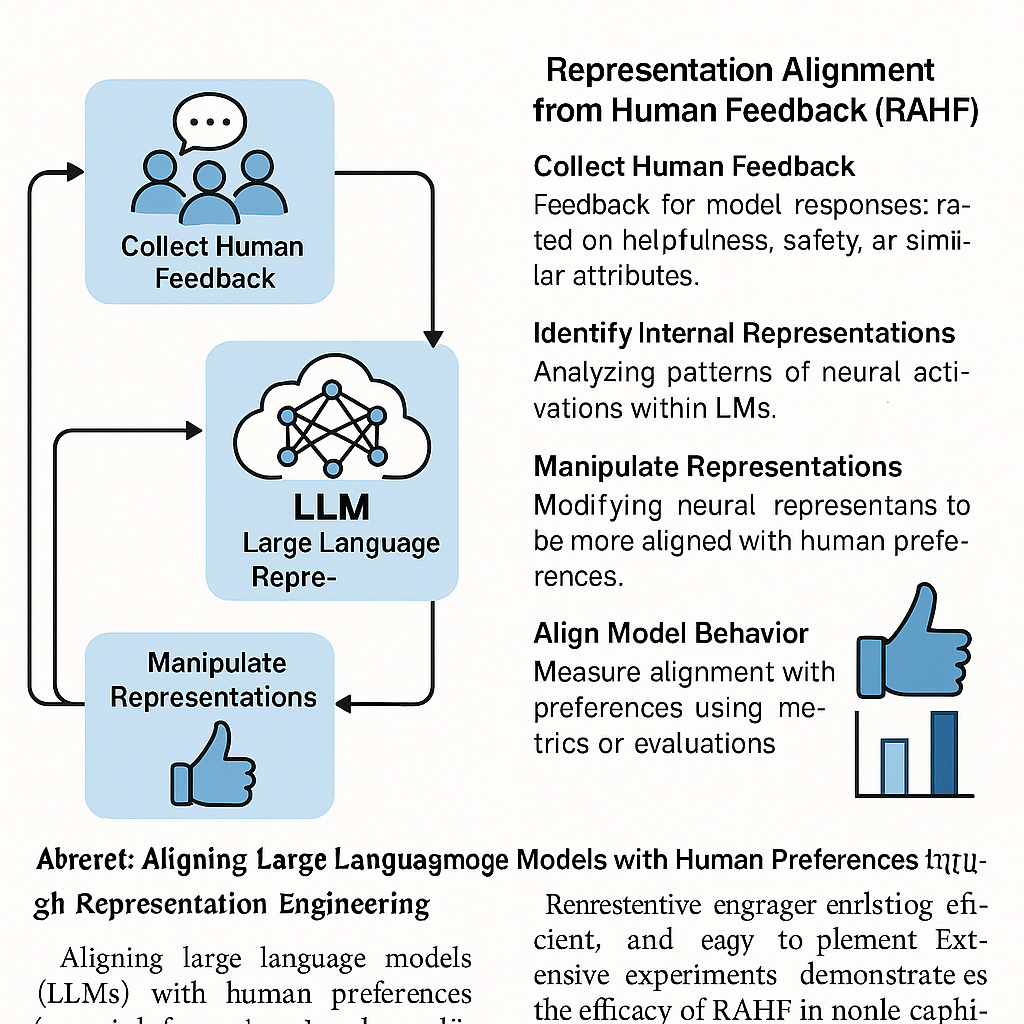}
        \caption{CoT}
        \label{fig:sub3}
    \end{subfigure}

    \begin{subfigure}[b]{0.3\textwidth}
        \centering
        \includegraphics[width=\linewidth,height = 4cm]{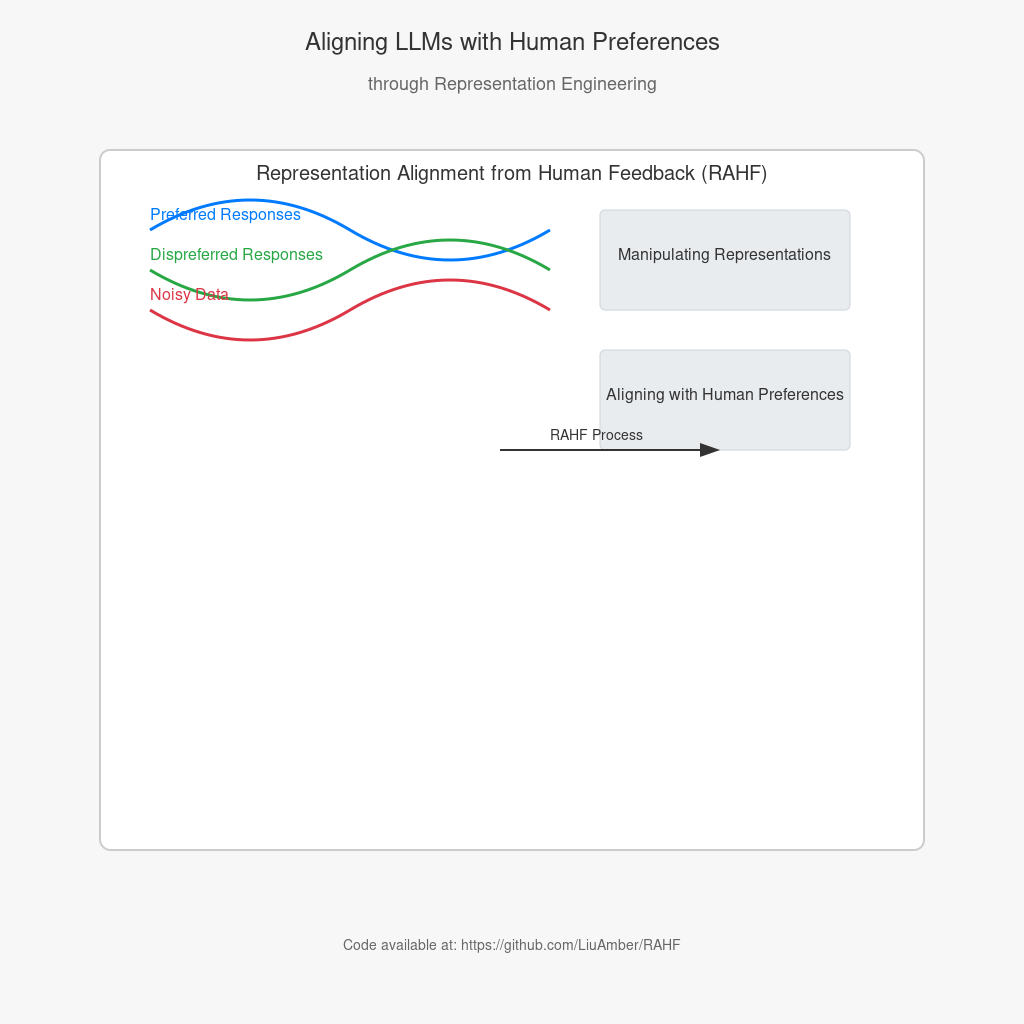}
        \caption{Zero-shot SVG}
        \label{fig:sub4}
    \end{subfigure}
    \hfill
    \begin{subfigure}[b]{0.3\textwidth}
        \centering
        \includegraphics[width=\linewidth,height = 4cm]{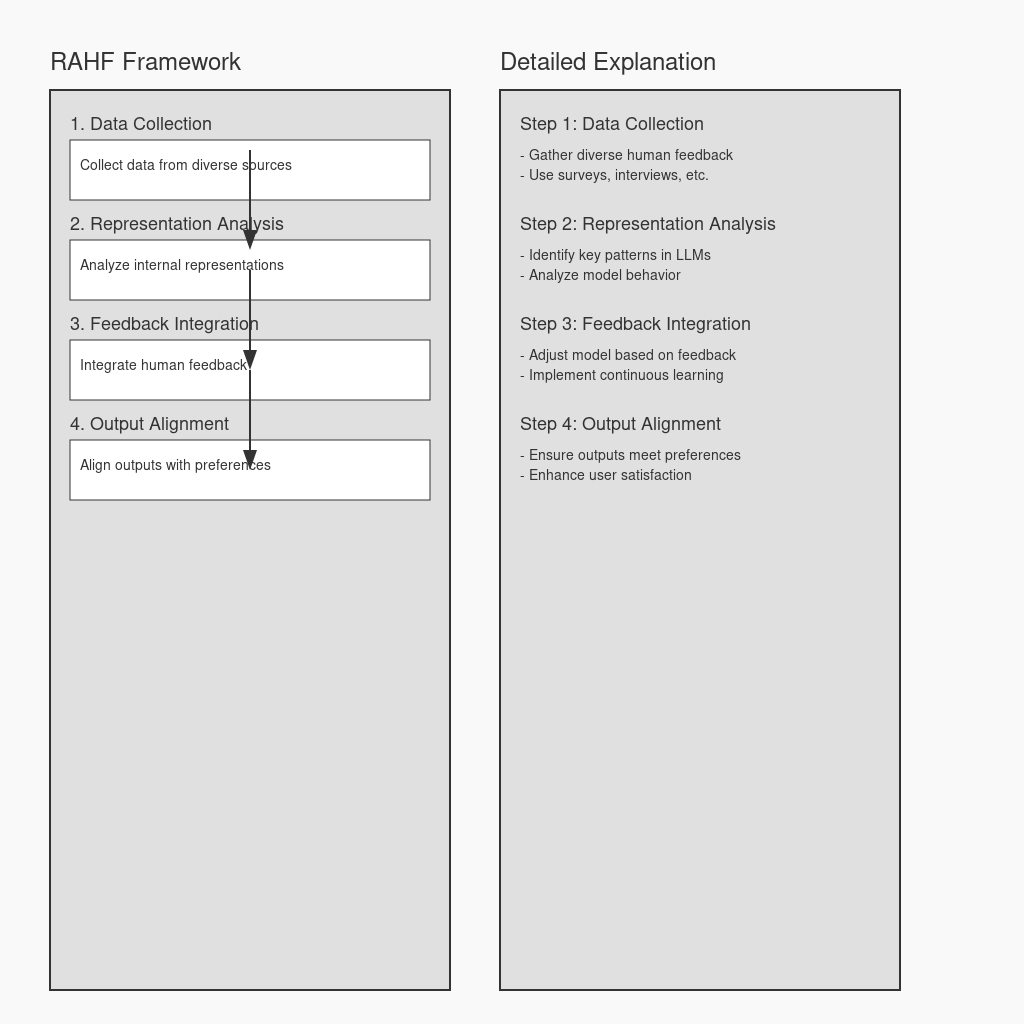}
        \caption{CoT SVG}
        \label{fig:sub5}
    \end{subfigure}
    \hfill
    \begin{subfigure}[b]{0.3\textwidth}
        \centering
        \includegraphics[width=\linewidth, height = 4cm]{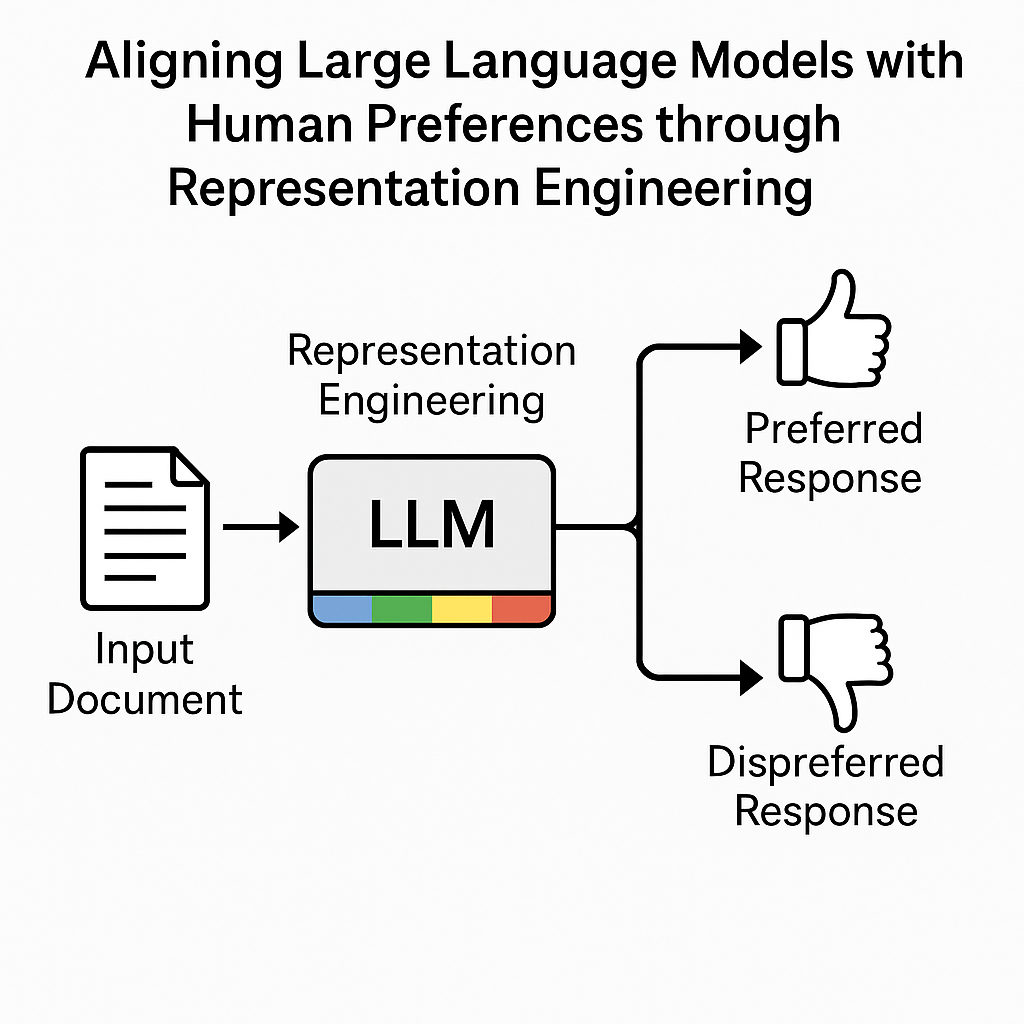}
        \caption{Chain-of-images}
        \label{fig:sub6}
    \end{subfigure}
    \caption{Figure 1 examples produced by both humans and models for all baselines from the paper \cite{Liu2023AligningLL}. In the first row are the ones from Humans, Zero-shot, CoT; the second row: Zero-shot SVG, CoT SVG, Chain-of-images. Where we can see that the ones from Zero-shot, CoT and Chain-of-images are relatively decent, although cropped.}
    \label{fig:baseline_example.png}
\end{figure*}

\paragraph{Textual content analysis}
Table \ref{tab:venue_lengths} in the Appendix reports the average lengths of key textual elements—title, abstract, introduction, and Figure~1 caption—for each venue. Titles in average have 9 words, while abstracts average 175 words. Introductions are much longer, averaging 648 words. Captions in CV and AI/ML venues (CVPR, ICLR, NeurIPS) are approximately 50\% longer than those in NLP venues (ACL, NAACL, EMNLP), averaging about 60 words versus about 40 in the latter.

\section{Evaluation and Analysis}
\label{sec:evaluation}
With \GenFig, we assess how effectively current VLMs generate \textit{Figure~1} for academic papers.

\begin{table*}[th]
  \centering
  \small
  \setlength{\tabcolsep}{3pt}%
  \renewcommand{\arraystretch}{1.15}%

  \resizebox{\textwidth}{!}{%
    \begin{tabular}{l c c c c c c c c c}
      \toprule
      \multirow{2}{*}{\shortstack{Approaches\\to construct \\figures}}
      & \multicolumn{6}{c}{VLM-as-a-Judge}
      & \multirow{2}{*}{\shortstack{Text-Rich\\Catastrophic\\Neglect Score}}
      & \multirow{2}{*}{\shortstack{DINOv2\\Score}}
      & \multirow{2}{*}{\shortstack{Human \\Preference\\MRR}} \\
      \cmidrule(lr){2-7}
      
      & Clarity & Faithfulness & \shortstack{Info. Density} & Interestingness
      & Legibility & Aesthetic & & & \\
      \midrule

      \rowcolor[HTML]{EFEFEF}
      \multicolumn{10}{l}{\textbf{Text-to-Image}} \\
      \addlinespace[0.3em]   
      Zero-shot        & 77.7 & 85.2 & 75.0 & 68.6 & 87.6 & 76.3 & 65.6 & 55.5 & 45.3\\
      CoT              & 70.1 & 79.5 & 68.6 & 61.7 & 83.8 & 70.9 & 52.4 & 51.4 & 40.1\\
      CoI  & 68.9 & 77.5 & 64.3 & 63.2 & 89.5 & 74.2 & 58.5 & 50.9 & 40.5\\
      \rowcolor[HTML]{EFEFEF}
      \multicolumn{10}{l}{\textbf{Text-to-SVG}} \\
      \addlinespace[0.3em]   
      
      Zero-shot SVG    & 39.8 & 65.2 & 31.5 & 33.2 & 87.0 & 53.6 & 35.9 & 40.1 & 19.6\\
      CoT SVG          & 42.7 & 61.1 & 35.4 & 29.7 & 84.3 & 46.0 & 32.7 & 41.6 & 20.8\\

      \midrule
      \rowcolor[HTML]{EFEFEF}
      Humans           & \textbf{87.4} & \textbf{99.5} & \textbf{86.2}
                       & \textbf{78.0} & \textbf{96.6} & \textbf{82.4}
                       & \textbf{94.2} & \textbf{100.0} & \textbf{78.7}\\
      \bottomrule
    \end{tabular}%
  }%
  \caption{%
    Comparison of the different baselines for generating Figure 1s (see \S\ref{sec:analysis} and \S\ref{sec:human}).
    All scores range from 0–100 (higher is better).
    As shown, image-generation approaches (e.g., CoI) achieve the highest scores, while text-based generation methods perform significantly worse.
    Overall, all automated methods fall well short of the performance of human-generated figures (final row), highlighting the difficulty of the task.
  }%
  \label{results}
\end{table*}
\vspace{\baselineskip}
\subsection{Baselines Systems}
\label{sec:baseline}

We introduce six baselines spanning both direct and multi-step generation, as well as text-to-image and text-to-SVG paradigms.

\paragraph{Zero-shot image generation:}
For the Zero-shot baseline, we prompt GPT-Image-1 to generate \textit{Figure 1} based only on the provided text description. 

\paragraph{Zero-shot SVG generation:}
This variant prompts GPT-4o mini to generate SVG code of an image, from the text description, offering a direct text-to-SVG alternative.


\paragraph{Chain-of-Thought (CoT) image generation:}
Inspired by \cite{wei2022chain}, this approach decomposes the generation into stages. GPT-4o mini is first prompted to identify the figure’s taxonomy, then propose a coarse layout (e.g., left-right, top-bottom). It next specifies content for each region of the layout. Finally, we combine the type, layout, region details, and original text description as input to GPT-Image-1, which generates the final figure.

\paragraph{Chain-of-Thought SVG generation (CoT SVG):}
This approach mirrors CoT but produces SVG code instead of an image. GPT-4o mini handles all steps, including SVG generation in the final stage.

\paragraph{Chain-of-Images (CoI):}
Chain-of-Images adopts a multi-step process similar to CoT, but focuses on region-wise generation. GPT-4o mini first determines the figure type and layout, then produces detailed descriptions for each region. Rather than generating the full figure in one step, GPT-Image-1 renders each region individually. The final figure is assembled by combining these region-level images.

\paragraph{Other models and failed runs:}
We attempted to run three additional methods: (1) a diffusion-based approach, (2) a text-to-SVG model, and (3) the text-to-TikZ method AutomaTikZ, proposed by Rodriguez et al.\cite{rodriguez2023figgentextscientificfigure, rodriguez2024starvectorgeneratingscalablevector} and Belouadi et al.\cite{belouadi2024automatikz}. However, the first two models are no longer publicly accessible, and AutomaTikZ failed to generate compilable TikZ code for any of the 120 test examples.

\paragraph{Example outputs:}
Figure~\ref{fig:baseline_example.png} has examples of human- and AI-generated \textit{Figure~1}s for the paper by \citet{Liu2023AligningLL}.

\begin{table*}[t]
    \centering
    \small
    \setlength{\tabcolsep}{3pt}%
  \resizebox{\textwidth}{!}{%
    \begin{tabular}{lccccccccc}
        \toprule
        \multirow{2}{*}{Metrics} 
            & \multicolumn{6}{c}{VLM-as-a-Judge} 
            & \multirow{2}{*}{\shortstack{Text-Rich Catastrophic \\ Neglect Score}}
            & \multirow{2}{*}{\shortstack{DINOv2\\Score}}
             \\
        \cmidrule(lr){2-7}
            & Clarity & Faithfulness & Info. Density & Interestingness & Legibility & Aesthetic &  \\
        \midrule

        Kendall’s $\tau$  & 0.68 & 0.59 & 0.71 & 0.68 & 0.30 & 0.63  & 0.55  & 0.36  \\
        Spearman's $\rho$    & 0.77 & 0.67 & 0.79 & 0.77 & 0.35 & 0.71 &  0.64& 0.44 \\
        
        \bottomrule
    \end{tabular}
    }
    \caption{The agreement between automatic metric with human preferences (discussed in \S\ref{sec:human}). 
    As can be seen, VLM-as-a-Judge metrics align most closely with humans on five of six aspects, peaking at \emph{Info. Density} ($\tau_b{=}0.71$, $\rho{=}0.79$), while other evaluation dimension show mild to fair agreement. 
    }

    \label{agreement}
\end{table*}


    
\subsection{Automatic Metrics}
\label{sec:metrics}


To better assess the quality of generated figures, we design several automated metrics that enable reproducible evaluation. Broadly, these metrics involve two categories: image-to-image comparison, which uses the reference image as the gold standard, and image-to-text comparison, which uses the accompanying text (figure caption) as the gold standard. Since these metrics are originally defined on different scales, we rescale all scores to a common range of [0, 100] for direct comparison.

\paragraph{Image-to-Image Comparison (DINOv2 Score):}
Following the DinoScore proposed in StarVector \cite{rodriguez2023starvector}, we adopt the DINOv2 Score, which leverages DINOv2 embeddings \cite{oquab2023dinov2} and cosine similarity to assess visual similarity between the generated figure and the golden (reference) figure.

\paragraph{Image-to-Text comparison (VLM-as-a-judge):}

We introduce a VLM-based evaluation metric that compares each generated figure to its corresponding caption, excluding the paper’s title, abstract, and introduction. The evaluation prompt defines six dimensions of figure quality:
(a) \textit{Clarity:} Does the figure clearly illustrate the paper’s core technical contribution, including key components and relationships?
(b) \textit{Faithfulness:} Does the figure accurately reflect the paper’s content without distortion or misrepresentation?
(c) \textit{Information Density:} Is the figure concise, avoiding clutter as well as excessive empty space?
(d) \textit{Legibility:} Is all text readable, with clear labels and no distracting artifacts?
(e) \textit{Interestingness:} Is the figure engaging and memorable, beyond being merely functional?
(f) \textit{Aesthetic Score:} Does the figure show visual balance, effective color use, and a coherent design?

For each aspect, GPT-4.1 assigns a numeric score from 0 to 10, with a brief justification. Scores are then rescaled to the [0,100] range. These dimensions collectively capture key qualities of effective figures, including conceptual reasoning (e.g., encoding core ideas and patterns), visual layout (e.g., spatial organization of elements), and aesthetic appeal (e.g., color harmony and visual coherence).

\paragraph{Image-to-Text comparison (Text-Rich Catastrophic Neglect Score):}
Inspired by the metrics proposed by \citet{grimal2024tiammetricevaluating}, we introduce a text-rich catastrophic neglect score. While Grimal et al. define catastrophic neglect as the omission or misrepresentation of one or more elements from the prompt, we extend this to evaluate whether core ideas from a figure caption are omitted or insufficiently represented in the generated figure.
This metric assesses the completeness of alignment between the figure and its caption. To operationalize it, we first extract a list of core ideas from each caption using an VLM. At evaluation time, another VLM is prompted to assess the extent to which each core idea is reflected in the generated figure. Each idea receives a score of 0 (not covered), 0.5 (partially covered), or 1 (fully covered). The final score is computed as the average across all core ideas for a given figure.

\subsection{Experimental Analysis}
\label{sec:analysis}

\paragraph{Text-to-Image outperform Text-to-SVG baselines:}

According to Table~\ref{agreement}, even the lowest-performing text-to-image setup (CoI) surpasses the best SVG-based method on \textit{Faithfulness}, \textit{Information Density}, and \textit{Interestingness}, suggesting image-generation models currently capture scientific-figure complexity better than code-driven SVG pipelines.

\paragraph{Chain-of-$x$ does not offer gains over Zero-shot baselines:}
To our surprise, Zero-shot baseline achieves the highest overall performance. 
This suggests that, despite the simplicity of its prompt strategy, Zero-shot generation benefits from the strong generalization capabilities of models like GPT-Image-1. More structured methods such as CoT and CoI underperform in comparison, indicating that step-wise decomposition does not yet yield clear benefits.

\paragraph{VLM-as-a-Judge reveals nuanced distinctions.}

The VLM-as-a-Judge scores reveal meaningful distinctions across dimensions. For example, CoT figures show modest gains in \textit{Information Density} and \textit{Interestingness} over CoI, but lower \textit{Faithfulness}. These differences suggest structured prompting may improve layout and content packing, but may also cause drifts from the input content.

\paragraph{Human-generated figures consistently outperform all automated methods:}

We observe this trend across all metrics. In particular, they achieve near-perfect \textit{Faithfulness} (99.5) and \textit{Legibility} scores. This underscores the difficulty of the figure generation and highlights the gap between human and machine capabilities in synthesizing clear, accurate, and engaging scientific visuals.

\paragraph{Possible Failure Reasons}
We observe that current models struggle mainly with spatial constraint satisfaction during generation (also shown in Figure \ref{fig:baseline_example.png}). Common failure modes—such as incorrectly positioned arrows and truncated text—reveal insufficient understanding of element dimensions, spatial relationships, and canvas size limitations, even when models successfully comprehend the underlying scientific concepts.

\section{Human Evaluation Study}
\label{sec:human}
To assess the robustness of our automatic evaluation and dataset quality, 
we conducted a user study involving NLP researchers serving as raters. 
These participants reviewed a mix of AI-generated and human-created \emph{Figure 1}s.

\paragraph{Setup}
Nine domain experts (each with at least one NLP or computer vision publication) evaluated a 106-paper test set.
For each paper, they were shown the title, abstract, introduction, and the \emph{Figure 1} caption, followed by six candidate \emph{Figure 1} images generated by different baseline methods.
The images were randomly shuffled to avoid position bias. Annotators ranked the six figures from best (1) to worst (6), with ties allowed. 
Each paper was evaluated by two or three annotators, yielding 636 ranked judgments.

\paragraph{Agreement among human annotators:}
We measure agreement between human annotators at the rank level using Kendall’s $\tau$ ($\tau_b$ variant which accounts for ties in the rankings). 
We also measured Spearman’s rank correlation coefficient, which measures the strength and direction of the monotonic relationship between two ranked lists. 
These metrics are applied to the full 1–6 ranked lists provided by each pair of annotators for a given paper.
Results indicate moderate agreement (Kendall’s $\tau = 0.50$, Spearman’s $\rho = 0.60$). Disagreements among humans likely stem from a blend of objective and subjective factors: ranking requires both technical judgment (e.g., faithfulness to content) and personal preference (e.g., aesthetics). Differences in what annotators prioritize---such as visual design versus information density---naturally reduce alignment.

\paragraph{Human evaluation of the baseline approaches:} 
We report the mean reciprocal rank (MRR) of each method in Table~\ref{results}. MRR is computed by taking the reciprocal of the rank assigned by human annotators to the target method’s figure (i.e., reciprocal rank$=\frac{1}{\text{rank}}$), and then averaging this value across all annotated papers. When multiple annotators are available for a paper, we average their  ranks (Borda count aggregation) before computing the reciprocal rank for that paper. 
As shown in the right-most column of Table~\ref{results}, human-generated figures (last row) are consistently preferred by annotators, with a substantial margin over all automated methods.

\paragraph{Agreement between humans and automatic metrics:}
To evaluate the reliability of our automatic metrics, we compute their agreement with human rankings using the same correlation measures as above. For papers with multiple human rankings, we apply Borda count aggregation to average ranks before comparison. Table~\ref{agreement} shows the results.
Among the metrics, VLM-as-a-Judge scores align most closely with expert judgments, particularly in assessing \textit{Information Density}---whether a figure is concise and free of clutter. However, it performs less well on \textit{Legibility}, likely due to limitations in recognizing text readability, spatial relationships, and labeling clarity. Similarly, \textit{DINOv2} shows a moderate agreement with human preferences.

\section{Conclusion}

We introduce \GenFig{}, a benchmark for evaluating generative AI models on their ability to produce scientific figures that are both visually coherent and conceptually accurate. 
We also introduce automatic metrics that shows strong agreement with human judgments, suggesting a more reliable path for assessing progress.
Despite recent advances in image generation, our evaluation of representative VLMs shows that current models fall short in generating figures with clear visuals and deep conceptual alignment---highlighting the challenges of cross-modal reasoning in scientific contexts.
By establishing this benchmark and dataset, we aim to drive progress in multimodal reasoning, visual abstraction, and scientific communication. We hope \GenFig{} encourages the development of models that help researchers convey complex ideas more clearly and intuitively, advancing the broader mission of AI for science.

\section{Limitations}

One limitation is that we focus on representative pipelines rather than exhaustive model comparisons. 
We use two publicly accessible representatives—GPT-4o mini and GPT-Image-1 (text-only vs. vision-grounded)—to provide a proof-of-concept benchmark and demonstrate the framework’s generality. Our focus for now is on contrasting figure-generation paradigms rather than exhaustively benchmarking all mainstream VLMs; extending to more recent state-of-the-art models is left for future work.

Another limitation is that we currently report only overall scores, without stratifying results by Figure~1 category or domain. We plan to add per-category and per-domain analyses in future work.

\bibliography{custom}

\clearpage
\appendix

\label{sec:appendix}

\begin{center}
    {\LARGE\textbf{Appendix}}
\end{center}

\section{\GenFig{} Dataset Statistics}
\label{Dataset Statistics}

In this section, we introduce the detailed statistics of our \GenFig{} Dataset.
Table~\ref{tab:venue_counts} summarizes the number of papers per venue after filtering, from 2020 to 2025. Data gaps (indicated by “—”) reflect years in which conferences were not held.
Table~\ref{tab:venue_filter_comparison} reports the raw and filtered paper counts, as well as filtering rates (\%) by venue and year. Filtering rates are generally stable across venues (80–90\%), though some fluctuations are observed (e.g., NeurIPS 2022: 67.0\%). This, in one aspect, demonstrates the consistent application of filtering criteria and supports the overall reliability of the benchmark data.

\begin{figure*}[ht]
    \centering
        \includegraphics[width=0.9\linewidth]{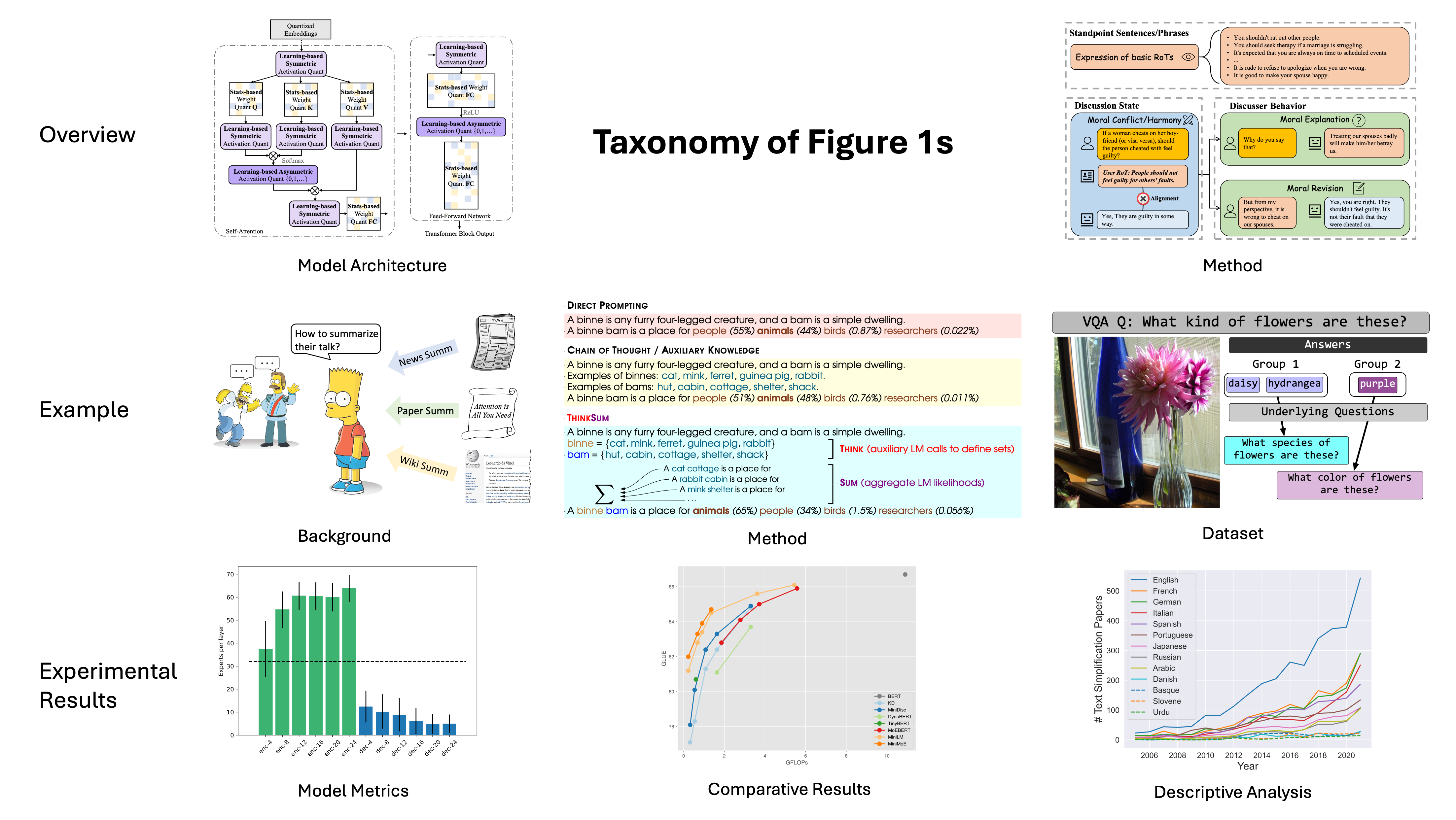}
    \caption{Examples for taxonomy of Figure 1s.}
    \label{fig:taxonomy.png}
\end{figure*}

As explained in the Taxonomy, Figure \ref{fig:taxonomy.png} shows representative examples illustrating the taxonomy of Figure 1s in scientific papers. 

\begin{figure*}[ht]
    \centering
    \begin{subfigure}[t]{0.48\textwidth}
        \centering
        \includegraphics[width=0.85\linewidth,]{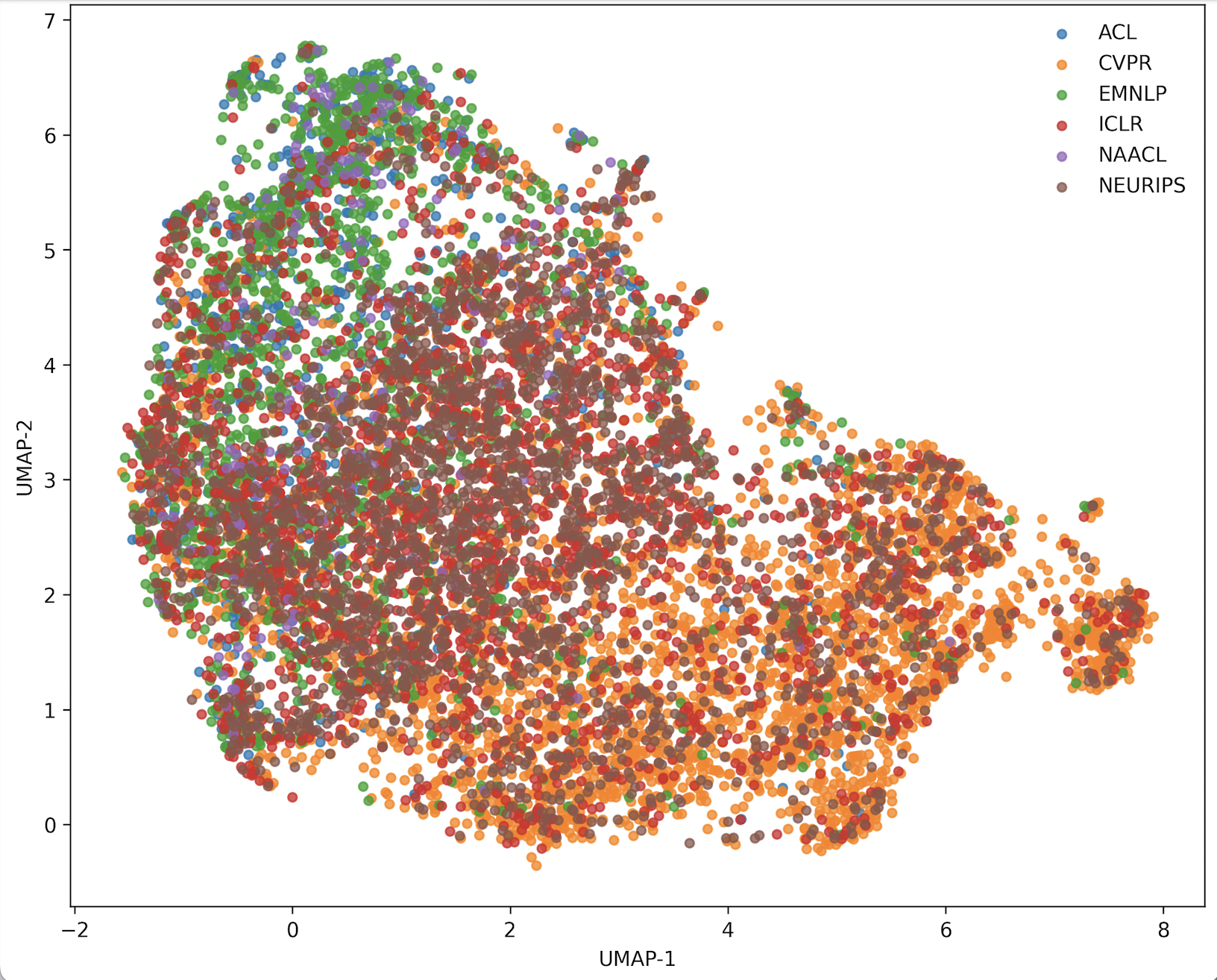}
        \caption{Embeddings clustered by venues}
        \label{fig:umap-venue}
    \end{subfigure}
    \hfill
    \begin{subfigure}[t]{0.48\textwidth}
        \centering
        \includegraphics[width=0.85\linewidth]{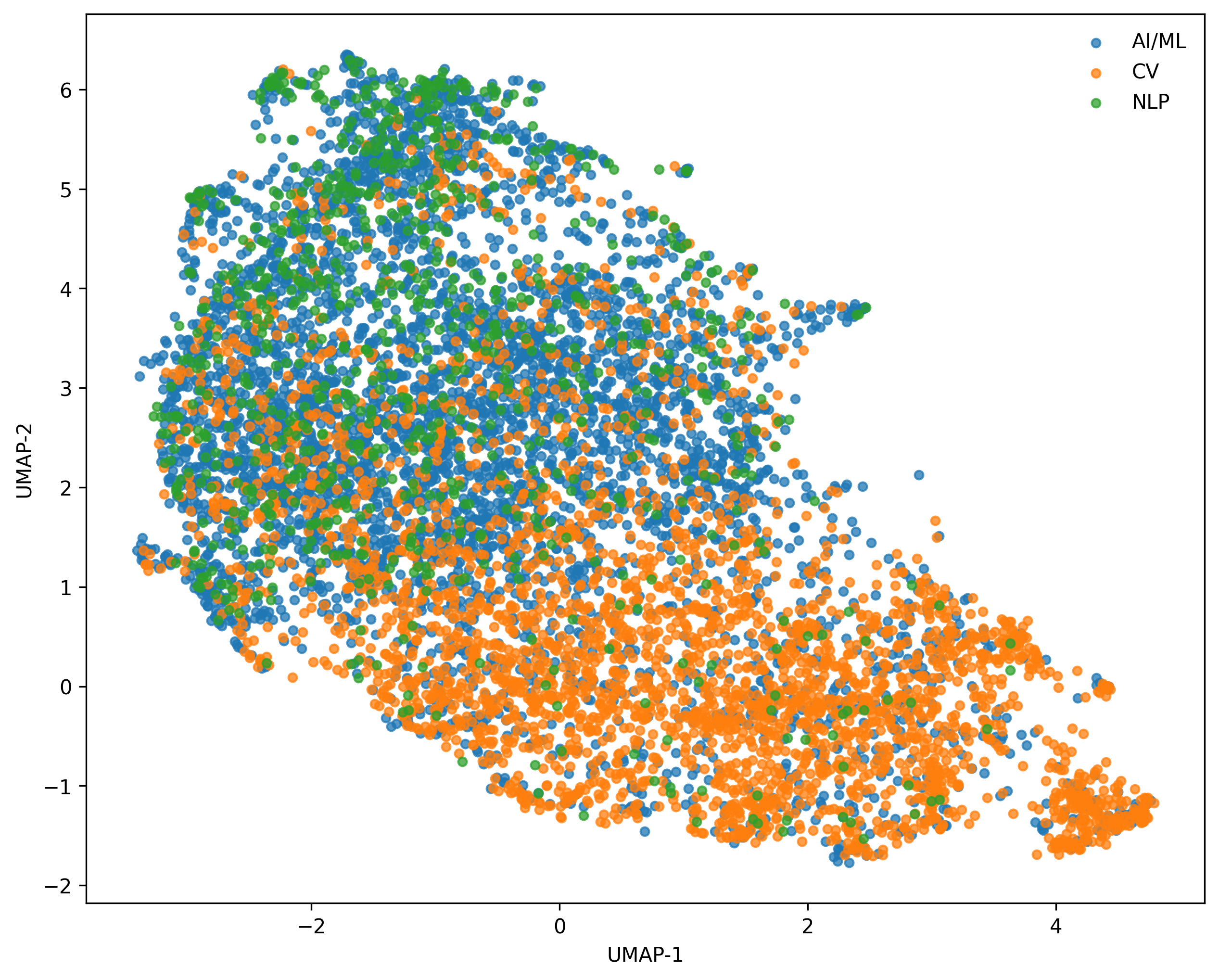}
        \caption{Embeddings clustered by fields}
        \label{fig:umap-field}
    \end{subfigure}
    \caption{UMAP visualizations of the paper representations: (a) clustered by venues and (b) clustered by research fields. Papers from different venues show some clustering tendencies, but there is considerable overlap, indicating shared or transferable representations across domains.}
    \label{fig:umap-comparison}
\end{figure*}

As discussed in \S\ref{subsec:data:analysis}, Table~\ref{tab:venue_lengths} summarizes average section lengths (title, abstract, introduction, caption) by venue. We can see that, CVPR, ICLR, and NeurIPS papers tend to have longer abstracts and captions than NLP venues. For figure analysis, CLIP-based visual embeddings were projected via UMAP; as shown in Figure~\ref{fig:umap-comparison}, CVPR figures form a relatively more distinct visual cluster, while figures from other venues and fields are less separable, which suggests a slight style difference across the venues and domains.

We also did the embedding space analysis across venues and research fields to try to inspect the differences between them. We use the CLIP ViT-B/32 model to extract 512-dimensional visual embeddings for all figures. These embeddings are projected into two dimensions using UMAP for visualization, enabling comparison of figure distributions across venues and research domains. Figure \ref{fig:umap-comparison} shows UMAP plots grouped by venue (a) and by research field (b). As we can see, figures from computer vision venues, such as CVPR, tend to form more coherent and distinct clusters in the embedding space. In contrast, figures from NLP and AI/ML venues (ACL, EMNLP, NAACL, NeurIPS, ICLR) exhibit substantial overlap. We could say, in general, Figure 1s from different venues and research fields tend to have more in common than not, which suggests that they might share similar visual styles and design conventions.

\begin{table}[ht]
  \small
  \setlength{\tabcolsep}{5pt}
  \centering
  \begin{tabular}{lrrrrrr}
    \toprule
    Venue   & 2020 & 2021 & 2022 & 2023 & 2024 & 2025 \\
    \midrule
    ACL      & 120  &  74  & 183  & 142  & 167  & —    \\
    NAACL    & —    &  54  &  69  & —    & 76    & —    \\
    EMNLP &  221 &  231  &  263 &  319 &  450 & —    \\
    CVPR     & 256  & 308  & 378  & 381  & 949  & 876  \\
    ICLR     & 116  & 159  & 199  & 308  & 498  & 841  \\
    
    NeurIPS  & 302  & 345  & 193  & 596  & 902  & —    \\
    \bottomrule
  \end{tabular}
  \caption{Number of papers per venue by year before and after filtering, covering the most recent five years (2020-2025).}
  \label{tab:venue_counts}
\end{table}

\begin{table*}[ht]
  \setlength{\tabcolsep}{3pt}

  \small
  \centering

  \begin{tabular}{lccc ccc ccc ccc ccc ccc}
    \toprule
    Venue
    & \multicolumn{3}{c}{2020}
    & \multicolumn{3}{c}{2021}
    & \multicolumn{3}{c}{2022}
    & \multicolumn{3}{c}{2023}
    & \multicolumn{3}{c}{2024}
    & \multicolumn{3}{c}{2025} \\
    \cmidrule(lr){2-4}\cmidrule(lr){5-7}\cmidrule(lr){8-10}\cmidrule(lr){11-13}\cmidrule(lr){14-16}\cmidrule(lr){17-19}
    & Raw & Filt & Rate
    & Raw & Filt & Rate
    & Raw & Filt & Rate
    & Raw & Filt & Rate
    & Raw & Filt & Rate
    & Raw & Filt & Rate \\
    \midrule
ACL     & 157 & 120 & 76.4 & 94 & 74 & 78.7 & 212 & 183 & 86.3  & 166 & 142 & 85.5 & 194 & 167 & 86.1 & — & — & — \\
NAACL   & — & — & — & 66 & 54 & 81.8 & 85 & 69 & 81.2 & — & — & — & 103 & 76 & 73.8 & — & — & — \\
EMNLP   & 259 & 221 & 85.3 & 268 & 231 & 86.2 & 312 & 263 & 84.3 & 397 & 319 & 80.4 & 530 & 450 & 84.9 & — & — & — \\
CVPR    & 285 & 256 & 89.8 & 333 & 308 & 92.5 & 424 & 378 & 89.2 & 428 & 381 & 89.0 & 1044 & 949 & 90.9 & 969 & 876 & 90.4 \\
ICLR    & 141 & 116 & 82.3 & 200 & 159 & 79.5 & 235 & 199 & 84.7 & 358 & 308 & 86.0 & 575 & 498 & 86.6 & 1010 & 841 & 83.3 \\
NeurIPS & 398 & 302 & 75.9 & 429 & 345 & 80.4 & 294 & 193 & 65.6 & 743 & 596 & 80.2 & 1091 & 902 & 82.7 & — & — & — \\
    \bottomrule
  \end{tabular}

    \caption{Comparison of paper counts before and after filtering, and retention rates (\%) by venue and year.}
  \label{tab:venue_filter_comparison}
\end{table*}

\begin{table*}[ht]

  \centering
  \small
  \begin{tabular}{l c *{4}{c@{/}c}}
    \toprule
    Venue & $n_{\text{papers}}$
    & \multicolumn{2}{c}{Title}
    & \multicolumn{2}{c}{Abstract}
    & \multicolumn{2}{c}{Intro}
    & \multicolumn{2}{c}{Caption} \\
    \cmidrule(lr){3-4} \cmidrule(lr){5-6} \cmidrule(lr){7-8} \cmidrule(lr){9-10}
  & & Char. & Words & Char. & Words & Char. & Words & Char. & Words \\
\midrule
ACL     & 686   & 68.3  & 9.3  & 963.3  & 157.1 & 3325.1 & 564.7 & 219.9 & 38.5 \\
NAACL   & 199   & 69.5  & 9.9  & 937.3  & 154.1 & 3223.0 & 551.4 & 223.1 & 39.9 \\
EMNLP   & 1484  & 70.3  & 9.7  & 971.1  & 158.2 & 3321.2 & 563.3 & 231.9 & 40.8 \\
CVPR    & 3148  & 65.7  & 8.9  & 1126.4 & 180.2 & 3969.5 & 666.2 & 331.3 & 56.5 \\
ICLR    & 2121  & 64.5  & 8.8  & 1140.6 & 183.5 & 3925.4 & 661.0 & 347.3 & 61.3 \\
NeurIPS & 2339  & 64.5  & 8.8  & 1118.0 & 181.0 & 4098.9 & 699.1 & 342.9 & 61.1 \\
\bottomrule
  \end{tabular}
    \caption{Average lengths (characters/words) for titles, abstracts, introductions, and Figure 1 captions, by venue. Across all 9,977 papers, a typical manuscript contains a 66-character ($\approx$9-word) title, 
a 1.09 k-character ($\approx$175-word) abstract, a 3.80 k-character ($\approx$648-word) introduction, 
and a 309-character ($\approx$54-word) Figure 1 caption. }
    \label{tab:venue_lengths}
\end{table*}

\begin{figure*}[p]
\centering
\begin{tcolorbox}[colback=gray!3!white, colframe=black!50, 
    title=Prompt for GPT-4.1 as a judge, width=0.98\textwidth]
\small
You are given a scientific paper's context (specifically the Fig.~1 caption) and the corresponding Fig.~1 image. Figure~1 typically serves important purposes in scientific papers: it may illustrate a key example explaining the paper's concept, introduce a pipeline or architecture of the method, or present data in other formats.

Your task is to evaluate how effectively and concisely the figure describes the underlying idea of the paper. Please provide insightful, objective, and constructive critiques based on established design principles and academic standards. Your evaluation should:

\begin{enumerate}
    \item Deliver comprehensive and unbiased assessments of the figure design.
    \item Identify potential areas for improvement with actionable feedback.
    \item Maintain consistent and high evaluation standards.
\end{enumerate}

Evaluation guidelines:
\begin{itemize}
    \item Score as objectively as possible.
    \item Apply rigorous grading standards where:
    \begin{itemize}
        \item 10 points = flawless figure,
        \item 7 points = mediocre figure,
        \item 4 points = figure with obvious shortcomings,
        \item 1--2 points = very poor figure.
    \end{itemize}
    \item Provide concise reasoning for each rating to avoid response truncation.
    \item Return \textbf{only} a valid JSON object with no additional commentary.
\end{itemize}

Grading criteria (score each from 0--10):
\begin{itemize}
    \item \textbf{Clarity:} Does the figure clearly demonstrate the paper's core technical idea? Does it effectively show input/output relationships or key components? 
    (10 = core concept is immediately clear; 1 = technical idea remains indecipherable.)
    \item \textbf{Faithfulness:} Does the figure avoid misleading statements or exaggerated claims not supported in the paper? 
    (10 = perfectly faithful; 1 = major misrepresentation.)
    \item \textbf{Information Density:} Does the figure use space efficiently without sparse or wasted areas? 
    (10 = optimally dense; 1 = excessive empty space or clutter.)
    \item \textbf{Interestingness:} Is the figure memorable and engaging? 
    (10 = highly compelling; 1 = generic and unmemorable.)
    \item \textbf{Aesthetic Score:} How well does the figure implement design principles (contrast, alignment, hierarchy, balance, modern styling)? 
    (10 = professional, clear design; 1 = poor design that hinders comprehension.)
    \item \textbf{Legibility:} Is all text readable at 100\% zoom? Are there typos, unclear annotations, or other impediments? 
    (10 = perfectly legible; 1 = completely unreadable.)
\end{itemize}

Return a JSON object with scores and brief justifications in this format:
\begin{verbatim}
{
  "Clarity": {"score": X, "reason": "Brief explanation"},
  "Faithfulness": {"score": X, "reason": "Brief explanation"},
  "Information Density": {"score": X, "reason": "Brief explanation"},
  "Interestingness": {"score": X, "reason": "Brief explanation"},
  "Aesthetic Score": {"score": X, "reason": "Brief explanation"},
  "Legibility": {"score": X, "reason": "Brief explanation"}
}
\end{verbatim}

\end{tcolorbox}
\caption{Prompt for GPT-4.1 as a judge} 
\label{app:prompt}
\end{figure*}

\begin{figure*}[p]
\centering

\begin{tcolorbox}[colback=gray!3!white, colframe=black!50, 
    title=Prompt for Text-Rich Catastrophic Neglect Score, width=0.98\textwidth]
\small
\textbf{Prompt to extract core concepts in the caption of the paper's Fig. 1}

User prompt is simply the caption and system prompt is as following:
\begin{verbatim}
The given input is the caption of a teaser figure in a paper. 
Extract the core ideas mentioned in this caption into a python list. 
Output the python list and nothing else.
\end{verbatim}

\textbf{Prompt to score the figure for Text-Rich Catastrophic Neglect Score}

User prompt is core concept list generated earlier and system prompt is as following:
\begin{verbatim}
The goal of the task is to evaluate how thoroughly and clearly 
the core concepts are conveyed in the input image. 
The core concepts are a list of concepts extracted from the teaser 
figure caption from this paper.      
The core concepts will be given in list format as string input 
and all of the concepts must be conveyed in the input image.
Use the paper information(paper title, abstract, introduction, 
teaser figure caption) provided below as reference and evaluate 
how thoroughly and clearly each concept is conveyed in the input image.

{paper_information}

Evaluate each core concept separately and output the score for 
each concept following the order of the concepts in the input list.
To convey a concept thoroughly and clearly means that the visual 
components and text in the image are both reasonable and necessary. 
The visual component and text that are conveying such concept 
need to be clear and meaningful.

Score can only take on the values 0, 0.5, 1
0 means that the concept was not covered in the image at all
0.5 means that the concept was partially covered in the image
1 means that the concept was thoroughly and clearly conveyed 
in the image
Provide the scores of each concept in a python list following 
the order the concepts are given in the input list.

Output only the scores list and nothing else. 
\end{verbatim}

\end{tcolorbox}
\caption{Prompt for Text-Rich Catastrophic Neglect Score}
\label{app:prompt_cata}
\end{figure*}



\section{Prompt and Additional Metrics}

The prompt used for VLM-as-a-judge (see Prompt~\ref{app:prompt}) is outlined as follows. It introduces the evaluation task, specifying that both the figure caption and corresponding image are provided as context. The prompt details the evaluation guidelines, clarifies the grading criteria across six dimensions, and defines the required JSON output format to ensure consistent and objective assessment. The prompts used to evaluate text-rich catastrophic neglect score (see Prompt~\ref{app:prompt_cata}) is outlined as following. There are two prompts that are used to evaluate text-rich catastrophic neglect score. The first prompt is used to extract the core concepts of the caption of Fig. 1 into a python list. The second prompt uses the core concept list to evaluate text-rich catastrophic neglect score based on how thoroughly the concepts are covered in the figure.

In addition to the metrics introduced in the paper, we also define two supplementary metrics: the Aesthetic Score Predictor and the Legibility Score (LS).

\paragraph{Aesthetic Predictor Score}
We computed aesthetic score given by a aesthetic score predictor as a reference. We chose \texttt{Aesthetic Predictor V2.5} as the predictor model because compared to \texttt{CLIP+MLP Aesthetic Score Predictor}, \texttt{Aesthetic Predictor V2.5} has been improved to evaluate wider range of images such as illustrations. While still very much different in style, illustrations are closer to scientific paper figures compared to natural images. Aesthetic score computed by \texttt{Aesthetic Predictor V2.5} is a floating point number between 1 and 10 and scores greater than 5.5 are considered good aesthetic scores according to the authors.

\paragraph{Legibility Score (LS).}
The Legibility Score (LS) measures the readability of text in a figure. We first apply OCR to localize text regions, then estimate the effective font size and the contrast between text and background for each region, following established accessibility standards. The overall LS is computed by aggregating the font size and contrast metrics across all detected text boxes, providing an objective evaluation of text legibility under standard viewing conditions.

As shown in Table~\ref{tab:merged_metrics}, the aesthetic and legibility scores do not always align with the expected quality of Figure1s. Notably, human-created Figure1s—which serve as the reference upper bound—sometimes receive lower scores than those assigned to figures generated by baseline methods.

\begin{table}[ht]

  \centering
  \resizebox{0.48\textwidth}{!}{
  \begin{tabular}{lccc}
    \toprule
    Baseline                 & Legibility Score     & Aesthetic Predictor Score  \\
    \midrule
    CoT                    & 0.80  & 3.27 \\
    Zero-shot SVG         & 0.34  & 3.30 \\
    CoT SVG               & 0.42  & 3.62 \\
    CoI            & 0.87  & 3.49 \\
    Zero-shot              & 0.86  & 3.35 \\
    \hline
    \rowcolor[HTML]{EFEFEF} Humans               & 0.75  & 3.41 \\
    \bottomrule
    
  \end{tabular}
  }
    \caption{Designed image embedding–based metrics}
  \label{tab:merged_metrics}

\end{table}

\end{document}